\documentclass{article}
\usepackage{spconf,amsmath,graphicx,hyperref}
\usepackage{cite}
\usepackage{amsmath,amssymb,amsfonts}
\usepackage{algorithmic}
\usepackage{graphicx}
\usepackage{textcomp}
\usepackage{xcolor}
\usepackage[raggedright]{sidecap}
\usepackage[singlelinecheck=false,justification=justified]{caption}
\usepackage{booktabs}
\usepackage{multirow}

\usepackage{caption}
\usepackage{subcaption}
\usepackage{enumitem}

\usepackage{sidecap}
\sidecaptionvpos{figure}{t}
\usepackage{balance}

\newcommand\blfootnote[1]{%
  \begingroup
  \renewcommand\thefootnote{}\footnote{#1}%
  \addtocounter{footnote}{-1}%
  \endgroup
}

\newif\ifworkinprogress
\workinprogresstrue

\ifworkinprogress
	\newcommand{\ms}[1]{\textcolor{blue}{{[Markus] #1}}}
	\newcommand{\mm}[1]{\textcolor{olive}{{[Marta] #1}}}
	\newcommand{\sn}[1]{\textcolor{green}{{[Shah] #1}}}
      
\else
    \newcommand{\ms}[1]{}
    \newcommand{\mm}[1]{}
    \newcommand{\sn}[1]{}
    
\fi

\makeatletter
\let\@oldmaketitle\@maketitle
\renewcommand{\@maketitle}{\@oldmaketitle
  \vspace{-8pt}
  \setcounter{figure}{0}
  \centering\includegraphics[width=0.95\linewidth]{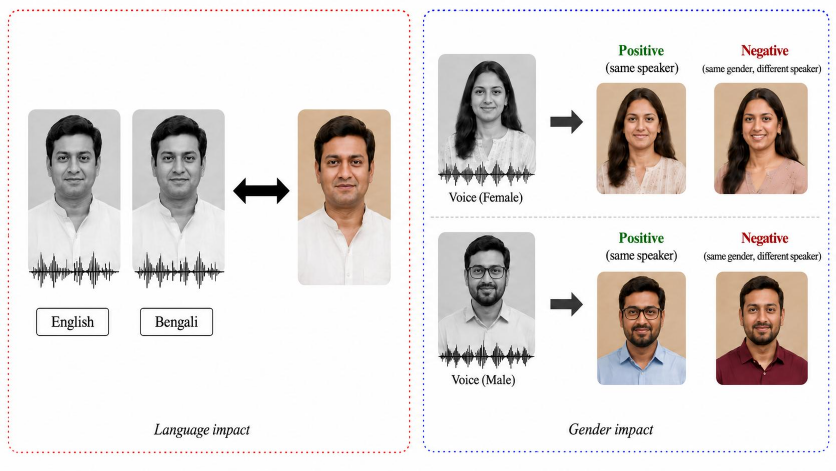}
  \vspace{-6pt}
  \captionof{figure}{Overview of the FLAG 2027 Grand Challenge. The aim is to evaluate face--voice association models across both language and gender dimensions. The ``language impact'' track focuses on whether a voice recorded in different languages can be correctly associated with the corresponding speaker's face. The ``gender impact'' track evaluates whether models show a performance deterioration when evaluated with negative-samples speakers of the same gender of the positive-sample speaker. The image is generated using ChatGPT to reflect same gender, different negative pairs. 
  }
  \label{fig:verification+protocol}
  \vspace{12pt}
 }
\makeatother

\def\BibTeX{{\rm B\kern-.05em{\sc i\kern-.025em b}\kern-.08em
    T\kern-.1667em\lower.7ex\hbox{E}\kern-.125emX}}

\makeatother

\title{\underline{F}ace-voice Association across \underline{LA}nguages and \underline{G}ender (FLAG) 2027 Challenge Evaluation Plan}

\name{
\begin{tabular}{c}
Marta Moscati$^{1}$\textsuperscript{\textdagger}, Swapnil Khandoker$^{1}$\textsuperscript{\textdagger}, Muhammad Saad Saeed$^{2}$\textsuperscript{\textdagger}, Shah Nawaz$^{1}$\textsuperscript{\textdagger},  \\ Fatima Noor$^{3}$, Rohan Kumar Das$^{4}$,  Mubashir Noman$^{5}$, Junaid Mir$^{3}$, \\Muhammad Haroon Yousaf$^{3}$, Khalid Malik$^{2}$, Markus Schedl$^{1,6}$  \\
\end{tabular}}
\address{
$^{1}$Johannes Kepler University, Linz, Austria, 
$^{2}$University of Michigan, Flint, USA,  \\
$^{3}$University of Engineering and Technology, Taxila, Pakistan,
$^{4}$Fortemedia Singapore, Singapore \\
$^{5}$Mohamed bin Zayed University of Artificial Intelligence, United Arab Emirates \\
$^{6}$Linz Institute of Technology, Austria \\
\tt mavceleb@gmail.com
}

\begin{document}
%
\maketitle
\begin{abstract}
Face--voice association models may rely on language or gender cues in the voice
rather than on speaker-specific voice characteristics, which can lead to a performance deterioration when the model has to identify a multilingual speaker or distinguis same-gender speakers.  To investigate these issues, we introduce the \underline{F}ace--voice Association across \underline{LA}nguages and \underline{G}ender (FLAG) 2027 Challenge. The challenge formulates face--voice association as a cross-modal verification task: given a voice, identify the speaker's face from a ``gallery'' of faces consisting of the speaker's face and a set of negative samples. Models are evaluated on identities not present in the training data (``unseen'') and both for languages present or absent from the training data (``heard'' and ``unheard''). Two evaluation settings are used to test models' reliance on gender: a  standard, unconstrained and a gender-constrained one, where the latter uses a same-gender gallery.
The performance of existing, baseline models in these settings reveals that models performance degrades under language shifts and in gender-constrained settings, highlighting the need to foster the development of models that capture identity-specific aspects beyond language and gender. The challenge provides a benchmark dataset, pretrained baseline models, and an evaluation framework to advance face--voice association.
\end{abstract}

\begin{NoHyper}
\blfootnote{\textsuperscript{\textdagger}Equal Contribution.}
\end{NoHyper}

\section{Introduction}
The face and voice of an individual contain distinctive characteristics and are widely used as biometric cues for person authentication, either independently or jointly in multimodal systems~\cite{jain2004introduction}. The strong perceptual correspondence that humans establish between faces and voices has motivated the development of automated face--voice association methods~\cite{nagrani2018learnable,nawaz2019deep,zheng2021adversarial,saeed2023single,hannan2025paeff,moscati2026face}. However, most existing studies evaluate the ability of automated systems to associate faces and voices only under controlled experimental settings that do not reflect real-world scenarios. Two often-neglected aspects are the possible change of language spoken by the speakers, and the model over-reliance on the speakers' demographic traits.

These neglected aspects pose serious limitations to real-world applications, since individuals may communicate in multiple languages and since an evaluation setting that is not realistic in terms of distribution of demographic traits, such as gender, can render verification results unrealistic. Given that a substantial proportion of the global population is bilingual or multilingual, it is important to understand whether face--voice association models remain reliable when the same speaker communicates across different languages. At the same time, gender-controlled evaluation is necessary to determine whether models rely on speaker-specific traits beyond the demographic ones, rather than simply leveraging the demographic differences of the speakers constituting the gallery of positive and negative labels
.

The \underline{F}ace--voice Association across \underline{LA}nguages and \underline{G}ender (FLAG) 2027 Challenge aims to study both these dimensions in two evaluation tracks: ``Language impact'' and ``Gender impact'', as summarized in Figure~\ref{fig:verification+protocol}. The challenge evaluates face--voice association models in multilingual conditions and introduces a gender-aware evaluation protocol, in which positive--negative pairs are formed from the ground-truth speaker and a speaker of the same gender. This design enables a more rigorous assessment of whether models identify speaker-specific face and voice traits that allow it to maintain good performance under cross-language and same-gender evaluations. 
The FLAG $2027$ Challenge has been accepted for support under the IEEE Signal Processing Society Challenge Program\footnote{\href{https://signalprocessingsociety.org/newsletter/2026/05/call-proposals-sps-challenge-program}{https://signalprocessingsociety.org/newsletter/2026/05/call-proposals-sps-challenge-program}}.


\section{Grand Challenge Objectives}
The goal of the FLAG 2027 Challenge is two-fold: 

\begin{itemize}[leftmargin=0.3cm]
\itemsep0em 
\item Evaluate face--voice association methods in multilingual conditions and avoiding models' reliance on gender as proxy for speakers' identities. This evaluation will result in an understanding of the impact of language and gender on the performance of face--voice association models.
\item Foster the development of multilingual and gender-aware methods that leverage speaker-specific traits instead of relying on demographic ones.
\end{itemize}

\begin{figure}[t]
    \centering
    \includegraphics[width=0.99\linewidth]{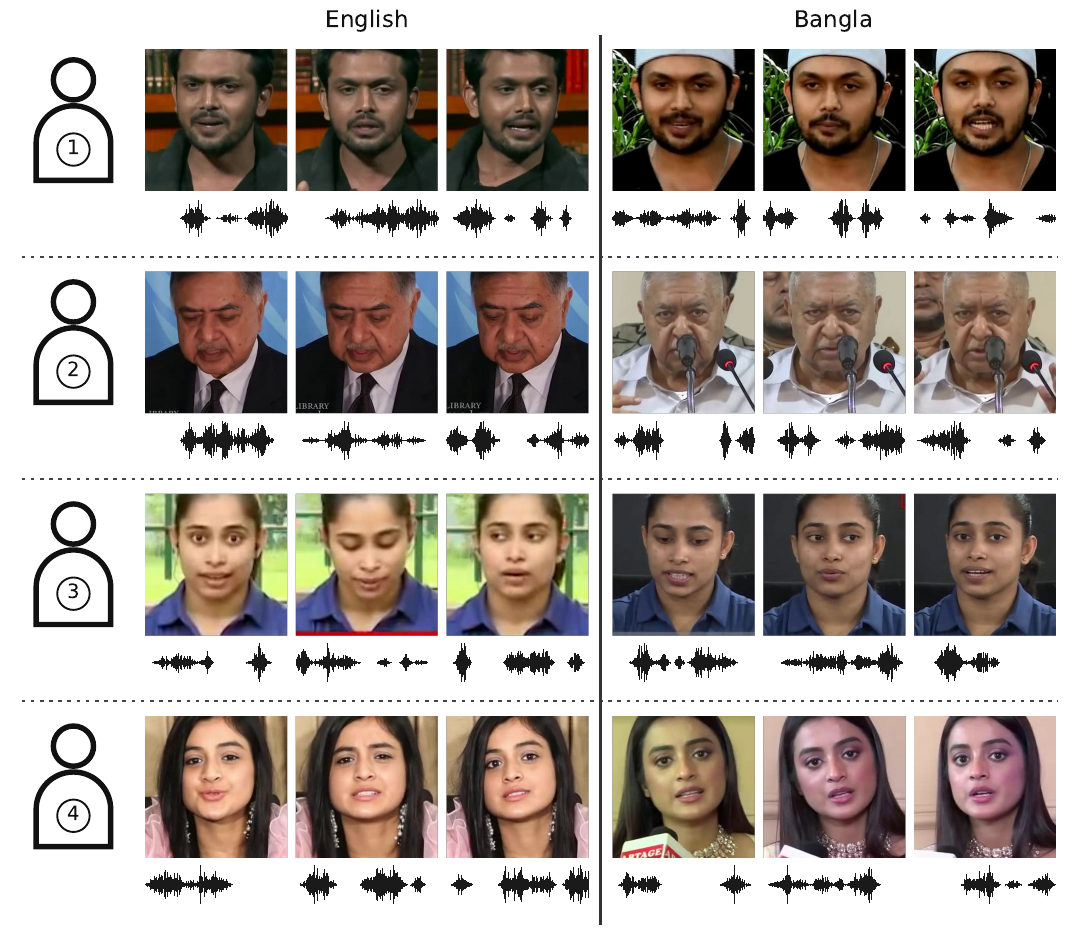}
    \caption{Audio-visual samples selected from the dataset. For a same speaker, left: English, right: Bengali. The visual data contains different variations such as pose, lighting condition, and motion.}
    \label{fig:mav_celeb_v3}
\end{figure}

\section{Grand Challenge Description}

\noindent \textbf{Dataset.} Building upon the prior FAME $2024$ \& $2026$ Grand Challenges hosted at ACM Multimedia~\cite{saeed2024synopsis} and IEEE International Conference on Acoustics, Speech, and Signal Processing~\cite{moscati2026linking}, we extended the MAV-Celeb dataset to include one additional language, curating a new dataset split consisting of $100$ English-Bengali speakers. The split provides language and gender annotations, which allow to analyze the impact of languages and gender on face--voice association. 
%
Table~\ref{tab:dataset_stats} summarizes the dataset characteristics. Each speaker appears in more than one video, for a total of 557 videos, and an average of 5.6 videos per speaker. Face--voice pairs are curated by sampling one video frame per second of active speaker, paired with the audio of the corresponding audio segment. 
The visual data spans a vast range of setups, including different poses, motion blurs, background clutters, video qualities, occlusions and lighting conditions. Moreover, since the videos originate from real-world situations, they reproduce the same challenges that are encountered when deploying face--voice association tools in real-world scenarios, such as noise, background chatter or music, overlapping voices, and compression artifacts. These aspects render the dataset both challenging for existing algorithms, and useful for developing algorithms that can have an impact on real applications. 
Figure~\ref{fig:mav_celeb_v3} displays audio-visual samples from the newly collected split.

The training, development, and evaluation splits of the datasets will be shared with the participants on the first day of the first phase of the challenge, i.e., the first day of the progress phase. These splits will be used by teams to develop their systems throughout the challenge duration. In order to allow participants to benchmark their results, in the progress phase we will also release a pretrained model based on recent research on face--voice association. Furthermore, we will allow teams to compare each others performance by hosting the challenge on on CodaBench\footnote{\href{https://www.codabench.org/competitions/18062/}{https://www.codabench.org/competitions/18062/}}.  
The input data required for the final evaluation phase  will be shared on the first day of the evaluation phase (without releasing ground-truth labels). This will ensure that the results of the evaluation phase are indicative of the generalization capability of the models developed during the progress phase.

\begin{table}[t]
\centering
\caption{Summary of the dataset characteristics.}
\begin{tabular}{lrr}
\toprule
\textbf{Dataset} & \textbf{Bengali} & \textbf{English} \\
\midrule
\# of celebrities                    & 100 & 100 \\
\# of male celebrities               & 57 & 57 \\
\# of female celebrities             & 43 & 43 \\
\# of videos                         & 301 & 256 \\
\# of hours                          & 41.6 & 29.9 \\
\# of utterances                     & 32{,}341 & 22{,}823 \\
Avg. \# of videos per celebrity       & 3.0 & 2.6\\
Avg. \# of utterances per celebrity   & 323.4 & 228.2 \\
Avg. length of utterances (in s)      & 4.6 & 4.7 \\
\bottomrule
\end{tabular}
\label{tab:dataset_stats}
\end{table}

\noindent \textbf{Baseline Method \& Starter Kit.}
To allow participants to benchmark results, we will release a pretrained instance of a novel competitive multimodal method for face--voice association~\cite{saeed2022fusion} . 
The model consists of a two-branch network that takes as input the embeddings of faces and voices. The embeddings to be used as input to the face-encoding branch are obtained using a well-established convolutional neural network pre-trained on a large-scale face recognition dataset~\cite{parkhi2015deep}.  The embeddings to be used as input to the voice-encoding branch are obtained using an audio encoding network for speaker recognition~\cite{xie2019utterance} trained using the language available in the training set (i.e., the \textit{heard} language). 
The multimodal model further combines the face and voice embeddings by projecting them into a shared space. The model is optimized by means of a loss function that imposes orthogonality constraints on the multimodal embeddings of different speakers.
We refer the readers to FOP~\cite{saeed2022fusion} and to the repository of the dataset\footnote{\href{https://github.com/SwapnilKhandoker101/FLAG_2027}{https://github.com/SwapnilKhandoker101/FLAG\_2027}} for more information on prior work on the baseline.

\begin{table*}[t]
\caption{Cross-modal verification between face and voice across various test configurations of MAV-Celeb V4 dataset. 
}
\label{tab:cross-modal}
\begin{center}
\begin{tabular}{lllccccc}
\toprule
& & & \multicolumn{2}{c}{Standard} &  \multicolumn{2}{c}{Gender-Constrained} & \multirow{2}{*}{Overall Score} \\
\cmidrule{4-7}
Method & Phase & Config. & Eng. test  & Bengali test  & Eng. test  & Bengali test &  \\
\midrule
\multirow{2}{*}{FOP~\cite{saeed2022fusion}} 
& Dev & Eng. train     & 32.54   & 38.12 & 	32.99   & 44.01  & 36.92 \\
& Eval & Eng. train    & 29.10   & 32.90 & 39.4     & 39.90  & 35.32 \\
\bottomrule
\end{tabular}
\end{center}
\end{table*}

\noindent \textbf{Baseline results.} Table~\ref{tab:cross-modal} provides the baseline face--voice association results  with the impact of languages and gender. The FLAG challenge $2027$ encourages participants to explore novel ideas to improve the performance of face--voice association models in cross-lingual and gender-constrained settings.

\noindent \textbf{Challenge Setup.} 
As described above, the dataset consists of videos of several speakers. Each video represents the speaker while speaking one language only. However, each speaker appears in videos of least two distinct languages. To test models' performance in a multilingual setting, the dataset is divided into train, development, and evaluation splits following the so-called \textit{unseen-unheard} configuration~\cite{nagrani2018learnable,nawaz2021cross}: the set of speakers of the train split is disjoint from the sets of speakers of the development and of the  evaluation splits, and within a split speakers speak the same language. 
Figure~\ref{fig:verification+protocol} shows the evaluation protocol at validation and test time. 
The evaluation is carried out on the task of cross-modal verification and on both a \textit{heard} language, i.e., available during training, and an \textit{unheard} language, i.e., \textit{not} available during training, to assess the impact of language. Gender-constrained positive and negative pairs are curated to evaluate the impact of gender.
Alongside the audios (.wav) and images (.jpg), the development and evaluation datasets will include the .txt files of the face--voice pairs, for which each line has the following format:
\begin{itemize}[leftmargin=0.3cm]
\itemsep0em 
    \item \texttt{ljAnhn41 English\_test/voices/00000.wav English\_test/faces/00000.jpg} 
    
 \dots
 
    \item \texttt{neHzLCeC English\_test/voices/00001.wav English\_test/faces/00001.jpg}
\end{itemize}
The first entry of the line represents the ID of the pair. The remaining two represent the local path of the corresponding audio, and of a visual frame extracted from the video. 
The dataset will be publicly available and provided alongside pre-extracted features representing the audios and images as encoded with state-of-the-art pre-trained architectures. 
We release all information related to the challenge on the challenge website.\footnote{\href{https://mavceleb.github.io/dataset/competition.html}{https://mavceleb.github.io/dataset/competition.html}}

\noindent \textbf{Evaluation Metric.} As commonly done for tasks of face--voice association~\cite{nagrani2018learnable,saeed2024synopsis,moscati2026linking}, we will evaluate models' performance with equal error rate (EER). The EER is the value of the false acceptance rate (FAR) and false rejection rate (FRR) for which both the errors are equal. As for FAR and FRR, a low value of EER indicates a good performance of the system. We expect participants to submit a .txt file containing output scores for every pair in the test set, indicating the system's confidence that the face and voice are matching, or in other words, that they belong to the same person. As we will indicate in the challenge description, for a same .txt submission file, a face--voice test pair having a higher score than another face--voice test pair will be interpreted as the model having a higher confidence that the first pair is \textit{matching}, i.e., that the face and voice correspond to the same person, as compared to the second pair.   
We choose this evaluation setup for several reasons. First, it is consistent with the evaluation of the related FAME $2024$ \& $2026$ Grand Challenges hosted at ACM Multimedia~\cite{saeed2024synopsis} and IEEE International Conference on Acoustics, Speech, and Signal Processing~\cite{moscati2026linking}. From a technical point of view, EER does not require the confidence scores of the models to span the same ranges in order to compare the model. The metric also does not require the use of a fixed threshold, as opposed to other metrics such as precision. Furthermore, in real-world applications, system developers may optimize the threshold on the confidence score to convert it to a binary value that determines the model's prediction on whether the face and voice belong to the same or to different person(s), depending on their specific needs. With a high threshold, the FAR is expected to be low, while the FRR is expected to be high. In summary, to evaluate the performance of different systems, EER is more suitable than threshold-dependent metrics such as accuracy, since it is independent of the threshold.
 Since models will be evaluated on four different tasks, each model will result in four different EERs. The overall score to decide on the challenge winners will be computed as $(\text {Sum of all EERs}) / 4$.

\noindent \textbf{Submission Format.} 
Participants must submit a ZIP archive containing one score file per protocol cell, placed in two folders named after the tracks. To create the archive, run zip -r submission.zip no\_gender gender from within the directory holding the two folders. The expected layout is:
\begin{itemize}[leftmargin=0.3cm]
\itemsep0em 
\item \texttt{no\_gender/sub\_score\_v4\_English\_heard.txt}
\item  \texttt{no\_gender/sub\_score\_v4\_Bangla\_unheard.txt}
\item  \texttt{gender/sub\_score\_v4\_English\_heard.txt}
\item \texttt{gender/sub\_score\_v4\_Bangla\_unheard.txt}
\end{itemize}
\noindent \textbf{Submission Platform.} 
The grand challenge will be implemented using CodaBench\footnote{\href{https://www.codabench.org/competitions/18062/}{https://www.codabench.org/competitions/18062/}}. Participants are expected to compute and submit text files including the \texttt{ID} and confidence scores in the following format:

\begin{itemize}
\item \texttt{ljAnhn41 1.162691}

\dots

\item \texttt{neHzLCeC 1.235319}
\end{itemize}
In the progress phase, each team will be allowed to submit a maximum of $150$ submissions, with a maximum $15$ per day. In the evaluation phase, the number of total submission will be limited to $15$.

\noindent \textbf{Rules for System Development.}
Since the FLAG 2027 challenge aims at analyzing whether face--voice association capabilities translate across language and gender constraints, we will enforce the following rules for participation:

\begin{itemize}[leftmargin=0.3cm]
\itemsep0em 
\item A pretrained encoder for faces or voices is allowed. 
\item The participants are required to submit a $2$ page system description in the ICASSP template to the challenge organizers. Teams without system description will be disqualified from the challenge. Teams describing a setup that violates one of the above rules will be disqualified. 
\item The participants are required to submit a link to a working version of their setup, e.g., on a platform for open-source development such as GitHub. Teams without code submission or with a setup that violates one of the above rules will be disqualified.
\end{itemize}

\noindent \textbf{{Tentative Timeline.}} The timeline is outlined below.

\begin{itemize}[leftmargin=0.3cm]
\itemsep0em 
\item Registration Period: 15 Sept.~-- 15 Oct.~2026
\item Progress Phase: 15 Sept.~-- 10 Nov.~2026
\item Evaluation Phase: 11 Nov.~--~18 Nov.~2026
\item Challenge Results: 23 Nov.~2026
\item Submission of System Descriptions: 27 Nov.~2026
\item Challenge Paper Submission: 10 Dec.~2026
\end{itemize}

\noindent \textbf{Funding.}
The FLAG $2027$ Challenge has been accepted for support under the IEEE Signal Processing Society Challenge Program.
Table~\ref{tab:prize_distribution} summarizes the prize allocation for the top five ranked teams.

\begin{table}[t]
    \centering
    \caption{Prize distribution among the top five ranked teams.}
    \begin{tabular}{lccccc}
        \toprule
        \textbf{Rank}  & \textbf{1st} & \textbf{2nd} & \textbf{3rd} & \textbf{4th} & \textbf{5th} \\
        \midrule
        \textbf{Prize} & \$2,000 & \$1,200 & \$800 & \$600 & \$400 \\
        \bottomrule
    \end{tabular}
        \label{tab:prize_distribution}
\end{table}

\section{Registration Process}
The following Google Form will be used to allow participants to register their teams to the challenge.
\href{https://docs.google.com/forms/d/e/1FAIpQLSeJH4uvcWNjqSi7CssNthIv80GdrszjIvuNp4UN77-KMNzgZg/viewform?usp=sharing&ouid=112056064502733681727}{Registration form}.

\section{Acknowledgments}
This research was funded in whole or in part by the Austrian Science Fund (FWF): Cluster of Excellence \href{https://www.bilateral-ai.net/home}{\textcolor{blue}{\textit{Bilateral Artificial Intelligence}}} (\url{https://doi.org/10.55776/COE12}) and the doc.funds.connect project \href{https://dfc.hcai.at/}{\textcolor{blue}{\textit{Human-Centered Artificial Intelligence}}} 

\bibliographystyle{IEEEbib}
\small
\balance
\bibliography{refs}

\end{document}